# Using Model-Theoretic Approaches to Uncover Linguistic Organization

Olivia Griffin[1] and Jerry Sun[2]

## I. Introduction

Various scholars have proposed the idea that there are different ways for a form-meaning pairing to be iconic, and that these different types of iconicity may interact with one another (Buchler, 1986; Reiger, 1998; Rozhansky, 2015). As a way of formalizing this idea, Lǐ and Ponsford (2018) identify five features pertaining to the form of fully reduplicated words that are in an iconic relationship with some aspect of a meaning that was found to be marked by total reduplication. Based on these formal features, they propose the following five dimensions of iconicity ('iconicities' in Lǐ and Ponsford (2018)) that can be manifested by reduplication patterns:

(1) Balinese Pluractional markers

| | |
|---|---|
| *keplug* | 'explode' |
| *keplug~keplug* | 'explode repeatedly' |
| *pa-keplug* | 'X (plural) explode simultaneously' |

(Arka and Dalrymple, 2017)

Notice that the repeated-explosion event is marked by a form that repeats keplug, while the event where all of the explosions happen at once (no repetition) is marked by a form that does not involve any repetition. Viewed through this lens, the Balinese pluractional prefix pa- is not entirely arbitrary, because it highlights the distinction between two types of pluractionality that are marked in Balinese. This is a case of **iconicity** because a property of the form (repetition or non-repetition) is also a property of the associated meaning.

In this paper, we consider pluractional markers in Kaqchikel, Karuk, and Yurok. Like Balinese, each of these languages marks one type of pluractionality via reduplication, and a different type of pluractionality via non-reduplicative affixation. Unlike in Balinese, however, the choice of marking strategy cannot be explained in terms of whether or not the meaning involves repetition. We argue instead that the way that pluractionality is marked in Kaqchikel, Karuk, and Yurok is iconic because reduplication is more **complex** that non-reduplicative affixation, and in these three languages, the more complex type of pluractionality is marked by reduplication, and the less complex type of pluractionality is marked by non-reduplicative affixation. This paper serves as a proof-of-concept for applying model-theoretic approaches to language as a lens that can help us to recognize linguistic organization that is not apparent on the surface.

[1] UCSD Linguistic, email: ogriffin@ucsd.edu
[2] University of Toronto Department of Mechanical and Industrial Engineering, email: jhsun@mail.toronto.ca

# II. Background

## A. Iconicity

In language, a (set of) form-meaning pairing(s) is iconic if at least one of the properties of the form is also a property of the associated meaning. An example from Haji-Abdolhosseini et al. (2002) of an iconic form-meaning pairing in Niuean is given in (2).

(2) Iconicity in Niuean

| | |
|---|---|
| *noko* | 'knock' (probably once but not necessarily) |
| *noko~noko* | 'knock many times' |

Various scholars have proposed the idea that there are different ways for a form-meaning pairing to be iconic, and that these different types of iconicity may interact with one another (Buchler, 1986; Reiger, 1998; Rozhansky, 2015). As a way of formalizing this idea, Lǐ and Ponsford (2018) identify five features pertaining to the form of fully reduplicated words that are in an iconic relationship with some aspect of a meaning that was found to be marked by total reduplication. Based on these formal features, they propose the following five dimensions of iconicity ('iconicities' in Lǐ and Ponsford (2018)) that can be manifested by reduplication patterns:

(3) Lǐ and Ponsford's (2018) iconicities

(a) **Identity:** the identical form of the base and the copy of it may reflect identical events, entities, etc.

(b) **Magnitude:** the increase in phonological bulk when the base is reduplicated may reflect an increase in the magnitude of the event, entity, or quantity that is expressed, or of a group of events or entities.

(c) **Discreteness:** the discreteness of the base and its copy may match discreteness in the events, entities, etc. that are denoted.

(d) **Proximity:** the adjacency of the base and its copy may reflect the close temporal proximity of events or spatial proximity of entities.

(e) **Sequentiality:** the uttering of the base and its copy in sequence may reflect the occurrence of events in sequence (limited to events).

Although Lǐ and Ponsford analyze iconic mappings of repetitive meanings as arising from an interaction between the dimensions of identity, magnitude, discreteness, and sequentiality, many authors (e.g. Reiger (1998), Stolz (2007)), treat repetition as a property of certain form-meanings pairings that is iconic in its own right. Therefore, the following dimension of iconicity can be added to the list in (3).

(4) Additional dimension of iconicity

**Multiplicity:** the multiplicity of copies may reflect the presence of multiple events or entities.

The dimensions of iconicity offer a framework for moving away from the oversimplified idea of iconicity as a binary property that stands in opposition to arbitrariness, and towards a more accurate view of iconicity as a gradient property. Under this view, (sets of) form-meaning pairings can be iconic in different ways, and one (set of) form-meaning pairing(s) can be more or less iconic with respect to one another (set of) pairing(s). Further, the dimensions of iconicity act as a lens that can help us to identify cases of iconic organization that may be harder to see with a less nuanced view of iconicity.

Lǐ and Ponsford's list does not include a dimension of iconicity that references a relationship between the complexity of the form and the complexity of the meaning. We will argue, however, that model-theoretic approaches offer a way to explore iconicity along the dimension of complexity, and that like the other dimensions of iconicity, the dimension of complexity can be helpful for identifying and understanding instances of linguistic organization. In particular, we will show how the pattern presented in the following section can be understood as iconic when the dimension of complexity is taken into account.

## B. Pluractionality

In this section, we present data from Kaqchikel, Karuk, and Yurok, each of which marks one type of pluractionality–event-internal pluractionality–via reduplication, and a second type of pluractionality–event-external pluractionality–via non-reduplicative affixation. As illustrated in the data, event-internal pluractionality corresponds roughly to a single event that involves some type of repetition, while event-external pluractionality corresponds to a set of multiple events of the same type. The distinction between the two types of pluractionality is discussed in more depth in section II.C.2.

The Kaqchikel peoples live primarily in the Midwestern Highlands of Guatemala, where the language is spoken by nearly 500 000 people (Richards, 2003). There are at least three pluractional suffixes in Kaqchikel (Henderson, 2012) but we focus here on the reduplicative suffix -Ca', which marks event-internal pluractionality, and on the non-reduplicative suffix -löj, which marks event-internal pluractionality.

In many of Henderson's examples, the suffix -löj is preceded by a vowel, which Henderson states is copied from the stem, and appears for phonological reasons. Because Henderson analyzes the vowel insertion as being separate from the process that marks pluractionality, we do not classify -löj as a reduplicative affix. The same vowel-insertion process applies before -Ca', except when the root to which the suffix attaches ends in velar fricative. -Ca' is classified as a reduplicative affix because the first consonant of the root is reduplicated to produce the onset of affix. The nucleus and coda of the root are fixed segments–a vowel and a glottal stop that do not change, regardless of the phonology of the root. The meaning distinctions between roots with the -löj suffix and roots with the -Ca' suffix is illustrated in (5). These examples are taken from Henderson (2012) and are reproduced faithfully except that I have underlined the verb root in each example, and have identified the pluractional suffixes with the gloss -PLRCT.

(5) Event-external and event-internal pluractionality in Kaqchikel

X-i-tzuy-e' *'I sat'*
COM-A1s-sit-P.ITV

X-i-tzuy-u**löj** *'I sat many times'*
COM-A1s-sit-**PLRCT**
[event-external pluractionality; non-reduplicative affixation]

X-in-Ø-tzuy-u**tzu'** *'I made the motion of sitting there repeatedly'*
COM-E1s-A3s-sit-**PLRCT**
[event-internal pluractionality; reduplication]

Speaker Comment: your bottom doesn't really hit the chair

The examples in (5) clearly show that -löj and -Ca' encode different types of repeated-event scenarios. While the use of reduplication to mark event-internal pluractionality is an example of iconicity along the dimension of multiplicity, the use of non-reduplicative affixation to mark event-external pluractionality does not appear to be iconic along any of the dimensions of iconicity introduced in the previous section. We will argue in section V, however, that systems that mark event-external pluractionality via non-reduplicative affixation and event-internal pluractionality via reduplication are iconic along the dimension of complexity.

The Yurok tribe is currently the largest tribe in California, with nearly 5,000 enrolled members. The language belongs to the Algic language family, and as of 2007, was spoken by only a few elders (Wood, 2007). The Karuk tribe has over 3, 700 enrolled members and 5, 000 registered descendents as of 2020, making it the second largest tribe in California. There are fewer than a dozen first language speakers of the language, however language revitalization projects have been established by tribal members and language activists.

Like Kaqchikel, both Karuk and Yurok mark event-internal pluractionality via reduplication, and event-external pluractionality via non-reduplicative affixation (Conathan and Wood, 2003). Karuk marks event-external pluractionality with the suffix -va, and event-internal pluractionality via total reduplication of the root. This is shown in (6). Summarizing Garret (2001), Wood and Conathan state that the Yurok reduplication pattern that marks event-internal pluractionality has the form CVCV- but will often surface as monosyllabic reduplication due to syncope. This is shown in (7), along with data illustrating event-external pluractionality in Yurok, which is marked with the infix -eg-.

(6) Event-external and event-internal pluractionality in Karuk

*ikxip* ‘(s.g.) to fly’
*ikxipí~**xipi*** ‘to flutter’

*ikremyáhiš(rih)* ‘to start to blow’
*Ikremyáhišrih-**va*** ‘to blow off and on’

(7) Event-external and event-internal pluractionality in Yurok

*menoot* ‘to pull’
*menoot~**menoot*** ‘to keep pulling’

*chyuuk’wen-* ‘to sit’
*chy-**eg**-uuk’wen-* ‘to sit often’

If, as we argue, the pattern presented in this section is iconic along the dimension of complexity, then the more complex marking strategy will be paired with the more complex meaning. We assess complexity from a computational perspective, and so it is necessary to have a computational model of the forms and meanings of verbs, as well as a computational representation of the processes that relate a (morphologically) marked form to an unmarked form, and a pluractional meaning to a non-pluractional meaning. In order for these representations to be as accurate as possible, they should align closely with the way that the processes are represented in linguistic work. To this end, the following section gives an overview of reduplication and pluractionality from a theoretical perspective.

# C. Reduplication and pluractionality

## 1. Reduplication

In reduplication patterns, part or all of a word is repeated, and the entire repeated form is paired with a meaning (which is often but not always different from the meaning of the unreduplicated form, if an unreduplicated form exists in the language). (8) gives an example from Niuean where a partially reduplicated verb is paired with a frequentative meaning (Sperlich, 1997), and (9) gives an example from Siwu where the fully reduplicated form of a verb is paired with an adjectival meaning (Dingemanse, 2015).

(8) Reduplication example: Niuean

*fakaava* ‘to dig a trench’
*fakaava~**ava*** ‘(frequentative) to keep digging trenches’

(9) Reduplication example: Siwu

*mini* ‘encircle’

*mini~mini* ‘round’

The repeated form of a word–called the reduplicated form–can be analyzed into two parts: the *base*, which can be conceptualized as the original non-repeated word, and the *reduplicant*, which can be conceptualized as the repeated portion (bolded in the examples above). As illustrated in (8) and (9), the reduplicant may be identical to the base, but it does not have to be. The reduplicant may be shorter than the base, or it may even have some segments that are not present in the base at all.

From the perspective of phonology, a primary goal of a good theory or model of reduplication is to explain how the phonological content of the reduplicated form is determined. This overarching goal subsumes many smaller questions. Three questions of particular importance in the context of this paper are (i) *where does the phonological content for the reduplicant come from?*, (ii) *what determines the base?*, and (iii) *what drives partial (as opposed to total) reduplication?*.

According to one widely-held view, reduplicants begin with no phonological content, and so the content of the reduplicant is filled in by referencing the phonological material in the base (Base-Reduplicant Correspondence Theory of reduplication, henceforth BRCT; McCarthy and Prince 1995, 1999). When reduplicants contain segments that were not present in the base (i.e. fixed segments), they are assumed to have been predetermined and inserted according to rules or requirements that are part of the grammar. Another popular view is that reduplicants begin with all of the phonological content that corresponds to the a particular morphological constituent (usually this constituent is the same as the base), and then may modify that content to meet phonological requirements in the grammar (Morphological Doubling Theory of reduplication, henceforth MDT; Inkelas and Zoll 2005). A common type of modification is truncation, where some of the phonological content is deleted resulting in a reduplicant that is shorter than the base (partial reduplication). Some other possible modifications are inserting additional segments, or changing existing segments.

Under the view that reduplicated forms are composed of two ‘parts’— the base and the reduplicant— the base is the original and the reduplicant is the copy. It is important to explain what determines the base because the phonological content in the base influences what phonological content is in the reduplicant. Often, the base is taken to be morphologically defined. In MDT, the base ‘part’ of the reduplicated form is one or more morphological units that may have been modified in order to meet grammatical requirements, and the input of the reduplicant is one or more morphological units that have the same semantic/syntactic features as the base, and can be modified to meet a separate set of requirements. In BRCT, the base is a unit of phonological material that is available to be (partially or completely) copied into the reduplicant. In both BRCT and MDT, the base is taken to be pre-determined by the grammar.

The base and the reduplicant tend to be very similar to one another in terms of their phonological content. Because of this, it is often assumed that there are grammatical requirements that are violated by mismatches between reduplicants and their bases (in the case of BRCT) or reduplicants and the phonological content that they began with (in the case of MDT). However, partial reduplication–where the reduplicant is shorter than the base, and so is not a perfect match–is very common. Therefore, when a theory of reduplication tries to explain how the phonological content of reduplicated forms is determined, it should also explain *what drives partial (as opposed to total) reduplication*. One approach to explaining the motivation behind partial reduplication is the idea of templates that restrict the length of the reduplicant (usually referencing prosody). The template is enforced by a 'size restrictor' which acts as a sort of reduplicant-specific grammatical requirement that obliges reduplicants to be a particular length. An alternative strategy, argued for by Spaelti (1997), is to take general grammatical requirements on the prosody of the language, and dictate the way that they interact with other grammatical requirements in such a way that they end up restricting the size of the reduplicant, but not the base. This strategy is distinct from the templatic approach because it does not depend on reduplicant-specific requirements that are intended to place size restrictions. Under both the templatic approach and the prosodic well-formedness approach, it is necessary to define a mechanism that restricts the size of the reduplicant in order to produce partial reduplication.

Although there are some different approaches to explaining where the phonological content of the reduplicant comes from, what determines the, and why we see partial reduplication, there is a good deal of convergence across theories as well. Our computational representation of reduplication aligns with BRCT in that the reduplicant is built via a process that copies phonological materia from the base. In line with both BRCT and MDT, the base is a unit of the unreduplicated form that is predetermined by the grammar, and partial reduplication is achieved via a built-in mechanism that limits how much of the base gets copied.

In the following section, we go into greater detail about the distinction between event-external pluractionality and event-internal pluractionality from a theoretical perspective. As with the theoretical discussion of reduplication, this overview serves as a basis for the computational representation of the two types of pluractionality.

## 2. Pluractionality

Within the Neo-Davidsonian approach to event semantics (Parsons 1990), verbs are objects that (i) describe the properties of events (e.g. the property of being a chopping event, for the verb *chop*), and (ii) define which individuals are involved in the event (e.g. the individual(s) who do the chopping, or the individual(s) that get chopped). Expressed more formally, verbs are conceptualized as predicates that (i) take events as arguments and describe their properties, and (ii) relate the event argument to individuals via secondary predicates that denote the thematic relation (e.g. *agent of*, *theme of*) between the event and the individual. If an event does not have the properties that the predicate describes (e.g. it is not a chopping event) then the event does not *satisfy* the predicate. A sentence is false if there is no event argument that satisfies the verbal predicate.

Many languages contain a class of predicates called pluractionals that cannot be satisfied in single event scenarios. In one of the seminal works on pluractionality, Cusic (1981) identifies three parameters along which a scenario can be pluralized: phase, event, and occasion. Building on this work, Mattiola (2019) formalizes the distinction as shown in (10).

(10) Parameters along which a scenario can be pluralized
  (a) *phase* refers to plurality within the scenario, e.g. *Isabella itched her arm* (several scratching motions form a single itching event.)

(b) *event* refers to plurality occurring on a single occasion, e.g. *Isabella itched her arm repeatedly* (the itching is performed repeatedly resulting in several itching events.)

(c) *occasion* refers to plurality displayed on several occasions, e.g. *Isabella itched her arms every time we entered the forest* (the itching is performed multiple times, but not repeatedly in the strictest sense. This is a plural event scenario, but not a repeated event scenario.)

Sentences that contain pluractionals may be satisfied by plurality along different parameters, but they are always false if there is only one event of the kind described by the predicate.

All pluractionals require a plural event argument, but different types of pluractionals may have specific requirements about which parameter needs to be pluralized. Henderson (2012) describes two types of pluractionals in Kaqchikel that are satisfied in different types of plural event scenarios. The first type of pluractional is satisfied by a plurality of events that can take place intermittently on different occasions and in different locations (e.g. *Naomi chopped down a tree occasionally*). The second type of pluractional is satisfied by the continuous repetition of (some subphase of) an event (e.g. *Naomi kept chopping at a tree*). The contrast is illustrated in (11) (repeated from (5)). In each example, the verb root is underlined, and the pluractional suffix (if present) is bolded.

The first example shows a non-pluractional or base verb, which does not require plurality along any parameter. The second example shows the pluractional form of the verb where the meaning is compatible with a plurality of events that take place intermittently. The third example gives the pluractional form of the verb where the meaning is compatible with repetition of a subphase of the event specified by the base verb.

(11) Event-external and event-internal pluractionality in Kaqchikel

X-i-tzuy-e' *'I sat'*
COM-A1s-sit-P.ITV

X-i-tzuy-u**löj** *'I sat many times'*
COM-A1s-sit-**PLRCT**
[event-external pluractionality; non-reduplicative affixation]

X-in-Ø-tzuy-u**tzu'** *'I made the motion of sitting there repeatedly'*
COM-E1s-A3s-sit-**PLRCT**
[event-internal pluractionality; reduplication]

Speaker Comment: your bottom doesn't really hit the chair

Henderson argues that the -löj suffix shown in (11) marks event-external pluractionality, while the reduplicative -Ca' suffix marks event-internal pluractionality. The distinction between external and internal pluractionality originates in Cusic (1981), and is discussed by others including Lasersohn (1995), and Wood (2007). Under Henderson's analysis, the distinction between external and internal pluractionals is based on whether the plurality of events form a group (in the case of event-internal pluractionality) or not (in the case of event-external pluractionality). Specifically, Henderson argues that event-external pluractionality is the eventive equivalent of bare plurals (e.g. trees), while event-internal pluractionality is the eventive equivalent of grove-type groups (where a plurality of individuals form the relevant type of group on the basis of their organization in space, e.g. a plurality of trees form a grove if they are close together and fill a bounded space of a certain size). This argument is based on the observation that just as bare plurals do not impose any constraints on how individuals are organized in space, event-external pluractions (henceforth EPs) do not place requirements on how events are organized in time. By contrast, grove-type groups do place restrictions on how individuals are organized in space, and event-internal pluractionals (henceforth IPs) likewise place restrictions on how events are organized in time (e.g. a plurality of chopping events would not satisfy an IP unless their temporal distribution is such that the events can be considered contiguous). Notice also that a grove can be conceptualized as an individual, while the word trees indicates a set of distinct individuals. Likewise, Henderson represents IPs as single events, and EPs as sets of events.

Another important difference between EPs and IPs is that sentences with event-external pluractionals entail minimally different sentences without the pluractional (e.g. Garnett ate breakfast regularly entails that Garnett ate breakfast), while the repeated events of IPs do not always have to satisfy the base predicate, and so sentences with event-internal pluractionals can fail entailment to minimally different sentences without the pluractional (e.g. Garnett nibbled (at) his breakfast does not entail that Garnett ate his breakfast). This asymmetric relationship is illustrated by the Kaqchikel sentences in (12) (repeated from (11)) , and is highlighted again by the sentence in (13):

(12) Failed entailment for IPs in Kaqchikel: sitl

X-i-<u>tzuy</u>-e' *'I sat'*
COM-A1s-sit-P.ITV

X-i-<u>tzuy</u>-u**löj** *'I sat many times'*
COM-A1s-sit-**PLRCT**
[event-external pluractionality; non-reduplicative affixation]

X-in-Ø-<u>tzuy</u>-u**tzu'** *'I made the motion of sitting there repeatedly'*
COM-E1s-A3s-sit-**PLRCT**
[event-internal pluractionality; reduplication]

Speaker Comment: your bottom doesn't really hit the chair

(13) Failed entailment for IPs in Kaqchikel: drink

X-Ø-in-<u>qum</u>-u**qa'** jun kaxlan ya' *'I kept drinking at the coke'*
COM-A3s-E3s-drink-**PLRCT** a foreign water
[event-internal pluractionality; reduplication]

Speaker comment: You don't finish it. You just keep making the motion.

It is impossible for the second sentence in (12) to be true and first to be false, therefore the sentence containing the EP entails the sentence containing the base verb. On the other hand, it is plausible that a person made the motion of sitting repeatedly but never actually sat, meaning that the last sentence in (12), which contains an IP does not entail the first sentence in (12). In (13), the speaker explicitly states that the sentence containing the base predicate is false in the scenario described by the sentence containing the IP.

The fact that the repeated events of IPs do not always have to satisfy the base predicate means that unlike with EPs, the domain of an IP (i.e. the set of events that can serve as arguments to the pluractional predicate) is not necessarily a subset of the domain of the base predicate. The domain of an IP is not a superset of the domain of the base predicate either, because there are single-event scenarios that satisfy the base predicate, but cannot satisfy the IP. Therefore, the domain of IPs and the base predicate from which they are derived overlap, but one cannot be directly obtained from the other. Of course, even the events that do not satisfy the base verb are not unrelated to those that do. For example, an event where a person makes the motion of sitting repeatedly but does not sit down is related to an event where a person sits. For the purposes of this paper, I formalize this relationship with the assumption that speakers can use world-knowledge about which events are implied by other events to determine whether a plurality of events constitutes repetition of a subphase of the type of event specified by the base verb.

This section has established that from a theoretical perspective, pluractionals are predicates (functions) that can only be satisfied by plural-event arguments. Henderson (2012) proposes that the property of being satisfied by non-atomic plural events (non-groups) versus atomic plural events (groups) is what underlies the opposition between event-external and event-internal pluractionals. In summarizing the cross-linguistic characteristics of IPs and EPs, Henderson also notes that while the plural events that satisfy EPs are guaranteed to also satisfy the base predicate, this is not always the case for the plural events that satisfy IPs. This observation is instantiated in the Kaqchikel data. The non-uniform behaviour with respect to entailment is not greatly emphasized in Henderson's work, and it seems that he takes it to be a consequence of the fact that the arguments of EPs are made up of other events, while the arguments of IPs are atomic events that are spatiotemporally superimposed over a plurality of events. However, we believe that the relationship between the set of events that satisfy the pluractional and the set of events that satisfy the base predicate–namely, the fact that the domain of EPs is a subset of the domain of the non-pluractional predicate from which they are derived, while the domain of IPs overlaps with but is not a subset of the domain of the non-pluractional predicate from which they are derived–is an important distinction in the way that the meanings of EPs and IPs are derived, and should not be ignored. Therefore, we centre the process of defining sets of events in our computational representation of how the meanings of EPs and IPs are calculated.

Having established some theoretical background for the processes whose complexity we will be considering, we now turn to the task of creating a formal model for both the forms and the meanings of verbs.

# III. Modeling the Forms and Meanings of Verbs

In this work, processes (reduplication, non-reduplicative affixation, deriving the semantics of IPs, and deriving the semantics of EPs) are represented as algorithms that take either a form (an unreduplicated/unaffixed form) or a meaning (a non-pluractional meaning) as their input, and give another form (a reduplicated/affixed form) or meaning (an IP/EP) as their output. Therefore, the forms and meanings of verbs need to be modeled in such a way that the information can be input to/output by algorithm. We have chosen to use sets as the formal structure that represents both the form and the meaning of verbs. In this section, we justify this representational choice and demonstrate how insights from formal linguistics can be translated into set notation.

## A. Modeling the form of verbs as sets

In current phonological theory, it is common to conceptualize phonological words as trees that can represent featural, segmental, and prosodic material. Focusing on syntax, Zwicky and Isard (1963) demonstrate that the notion of trees found in linguistic theory corresponds to a particular type of formally defined tree–a labeled, ordered, finite, directed, singly-rooted, connected graph without circuits–and discuss how these formal objects can be captured with set theoretic notation. Here we follow their method, showing how the same type of formal structure can be used to represent phonological trees, and illustrating how the form of a verb can be translated into set theoretic notation.

A simplified representation of the structure of the English phonological word [stɪk] (stick) is given in Figure 1. This representation shows four segments, and the ordering of the segments with respect to one another. It also shows their organization into the onset, nucleus, coda, and rhyme of a syllable, and the fact that the syllable belongs to a prosodic word.

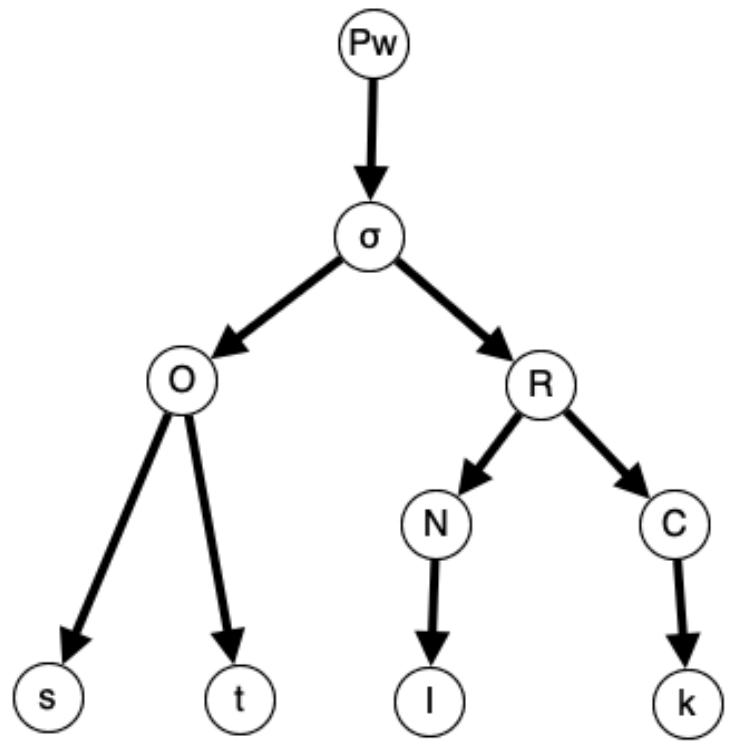

Fig. 1. Representation of the structure of [stɪk]

The same information can be conveyed by the structure in figure 2 along with a set, L, of labels for the vertices (circled numbers) in the tree, and a function, N, that dictates which label each vertex receives. Note that the function can be written as a set of ordered pairs where the first element (or input to the function) is a vertex in the tree, and the second element (or output of the function) is the label that gets assigned to the vertex.

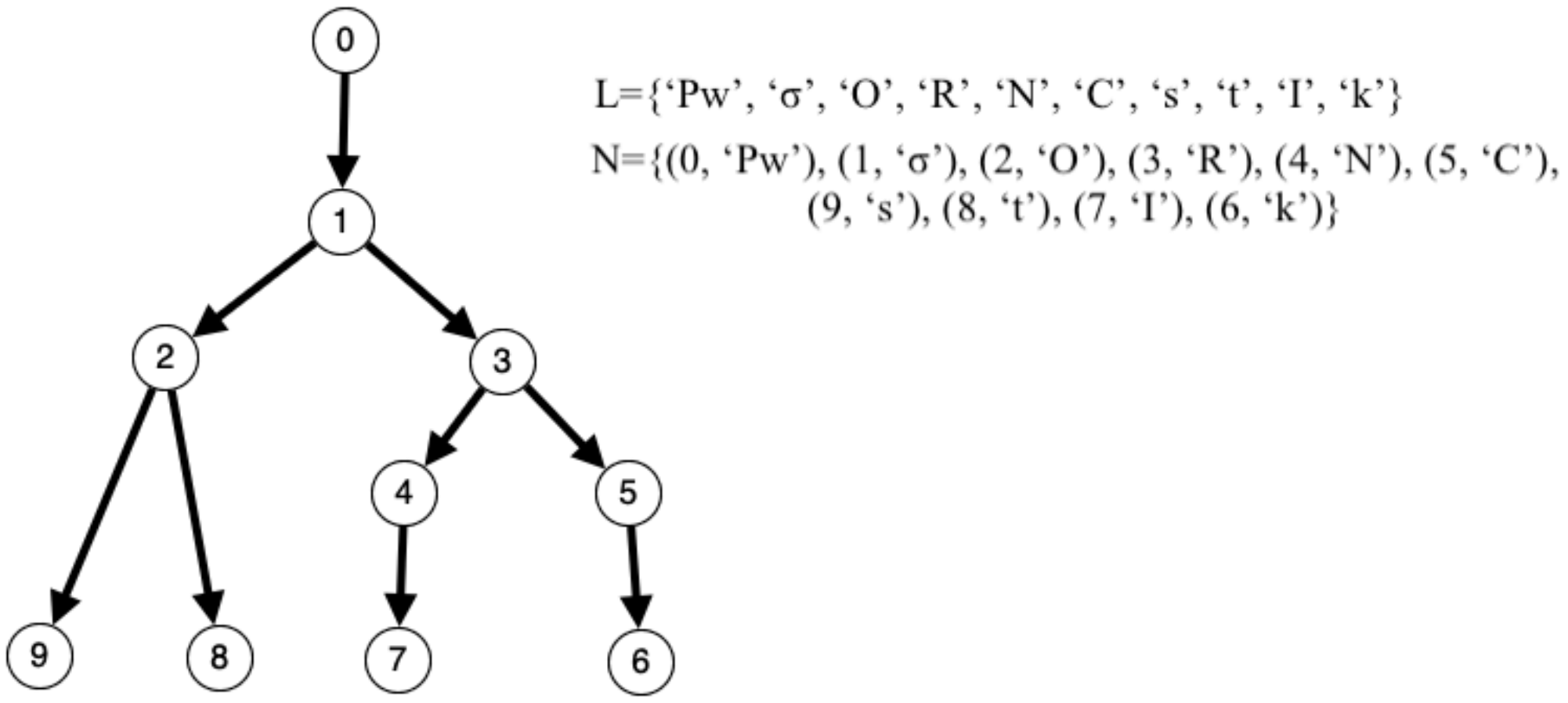

Fig. 2. Updated representation of the structure of [stɪk]

In order to completely represent the phonological structure in figure 1 in terms of sets, it is necessary to translate the information captured by the structure in figure 2 into sets as well. The vertices in that structure provide information about which phonological objects (segments, prosodic units, etc.) make up the word [stɪk]. The edges in the structure (the arrows that connect the vertices) provide information about the hierarchical organization of vertices. The edges show how the vertices are grouped–for example, the fact that vertices 9 (`s') and 8 (`t') are dominated by the same vertex, but vertex 7 ('ɪ') is dominated by a different vertex, i.e. belongs to a different prosodic unit. In addition, the direction of the arrows indicates that vertex 0 (`Pw') dominates all of the other vertices, rather than being dominated by all of the others. Finally, the left-to-right ordering of the vertices provides information about their precedence relationships, e.g. the fact that [s] precedes [t] and that the nucleus precedes the coda. These three pieces of information–vertices, dominance relations, and precedence relations–are encoded in the sets V, D, and Pr respectively. Note that following the approach laid out in Zwicky and Isard (1963), precedence is only defined for vertices that are dominated by the same vertex, and dominance is only defined for vertices between which there is no other intervening vertex. However, the precedence and dominance relations that are not directly defined in B and D can be determined via transitivity. In addition to the sets already listed, it is necessary to define a set U which stands for the universe of all objects. U contains all of the objects (vertices and labels) associated with the structure. Note that the edges in the structure in figure 2 are not included in U because these are represented in D as relations between vertices, not as independent objects. Therefore, the full translation of the phonological tree in figure 1 into a set theoretic representation is given in (14).

(14) Set-theoretic representation of the structure of [stɪk]

$U = \{V \cup L\}$
$V = \{0, 1, 2, 3, 4, 5, 6, 7, 8, 9\}$

$L = \{'Pw', \text{'σ'}, 'O', 'R', 'N', 'C', 's', 't', \text{'ɪ'}, 'k'\}$
$D = \{(0, 1), (1, 2), (1, 3), (2, 9), (2, 8), (3, 4), (3, 5), (4, 7), (5, 6)\}$
$P = \{(2, 3), (4, 5), (9, 8)\}$
$N = \{(0, 'Pw'), (1, \text{'σ'}), (2, 'O'), (3, 'R'), (4, 'N'), (5, 'C'), (9, 's'), (8, 't'), (7, \text{'ɪ'}), (6, 'k')\}$

This model shows which linguistic objects are present, and how they are organized with respect to one another. Because sets are well-defined formal structures, when the form of a verb is modeled in this way, it can serve as the input or the output of an algorithm. Therefore, the model meets our goals of capturing linguistic insights and translating them into a formal representation. We now turn to modeling the meaning of verbs as sets, keeping the same goals in mind.

## B. Modeling the meaning of verbs as sets

In the framework of neo-Davidsonian event semantics, verbs are represented as predicates that (i) take events as arguments and describe their properties, and (ii) relate the event argument to individuals via secondary predicates that denote the thematic relation between the event and the individual. An example of the denotation of the verb kiss (as it would be represented in formal semantics) is given in (15).

(15) Formal semantic representation of the meaning of the verb kiss

$$[[kiss]] = \lambda y. \lambda x. \lambda e. [kiss(e) \wedge (ag(e) = x) \wedge (th(e) = y)]$$

This denotation encodes 3 pieces of information which combine to give the meaning of the verb: (i) the number and type of arguments that the verb requires. In this case, the verb takes one event and two individuals as its arguments. (ii) restrictions that define which elements from the domain of events may serve as the event argument of the verb. In this case, only events which have the property of being an event of kissing are acceptable arguments. (iii) the relationship between the arguments. In this case, one individual must be the agent of the event, and the other individual must be the theme of the event.

The same three pieces of information can be represented using a set of ordered sets like in (15), where property_v($D_\epsilon$) is a function that takes as its input the domain of events and gives as its output the set of events that have the property dictated by the verb (e.g. the property of being a kissing event), agent($D_e$) is a function that takes as its input the domain of individuals and gives as its output the set of individuals that have the requisite properties to be the agent of a verb (e.g. animacy), and theme($D_e$) as its output the set of individuals that have the requisite properties to be the theme of a verb. Note that in section V, we will define property() as a class of functions with a specific instantiation for each verb. We use a generic function called property_v($D_\epsilon$) in (15) as a shorthand for ‘some function from the class of property() functions’. With this representation, the elements of each set give the number of arguments, the order of the elements encodes the relationship between the arguments of the verb, and the set inclusion requirement $e \in$ property_v($D_\epsilon$) gives the restrictions that define which elements from the domain of events may serve as the event argument of the verb. Note that in the set representation, the functions agent(x) and theme(x) define sets of individuals that may serve as the agent or theme of the verb, rather than establishing the relationship between the arguments as is the case with the function ag(e) and th(e) in (15). A general representation of this model of verb meanings is given in (16), and (17) gives the model for the verb kiss.

(16) General model for the meaning of verbs

[[verb]] = $\{\{e_1, x_{11}, x_{12}, \ldots\}, \{e_2, x_{21}, x_{22}, \ldots\}, \ldots, \{e_m, x_{m1}, x_{m2}, \ldots\}\}$ where, for all $i \in \{1, \ldots, m\}$, $e_i \in$ property_v($D_e$), and for all $j \in \{1, \ldots, m\}$ and $k \in N$, $x_{jk} \in$ agent($D_e$) or theme($D_e$). In other words, this is the set of all sets that meet the requirements for the verb.

(17) Model for the meaning of kiss

[[kiss]] = $\{\{e_1, x_{11}, x_{12}, \ldots\}, \{e_2, x_{21}, x_{22}, \ldots\}, \ldots, \{e_m, x_{m1}, x_{m2}, \ldots\}\}$ where, for all $i \in \{1, \ldots, m\}$, $e_i \in$ property_kiss($D_e$), and for all $j \in \{1, \ldots, m\}$ and $k \in N$, $x_{jk} \in$ agent($D_e$) or theme($D_e$). In other words, this is the set of all sets that meet the requirements for the verb.

When modeling the form of verbs, it is necessary to represent the linguistic objects and their organization with respect to each other. When modeling the meaning of verbs, it is necessary to show what type of event can satisfy the verb as well as the individuals (agents, themes, etc.) that can be arguments of the verb (if applicable). In section IV, we have shown how this information can be encoded in a set-theoretic model of the forms and meanings of verbs. These set-theoretic models serve as input to the algorithms that represent reduplication, non-reduplicative affixation, the process of deriving the meaning of IPs, and the process of deriving the meaning of EPs. In the following section, we now turn to defining these algorithms and determining their complexity.

# IV. Measuring Complexity

## A. Algorithms

In this section, we detail the steps required to derive the model of an affixed or reduplicated verb from the model of the unaffixed verb, as well as the steps required to derive the model of meaning of an EP or IP from the model of the meaning of the base verb. These series of steps are the algorithms whose computational complexity we calculate in subsection B in order to attach a measure of computational complexity to the processes of non-reduplicative affixation and reduplication, and to the derivation of the meanings of EPs and IPs.

# 1. Modeling the derivation of the form of pluractional verbs

In the input for each verb, there is an ordered set W that contains all of the information about the form of the base (i.e. non-pluractional) verb. The first element of W is another set, P, which contains information about the verb's phonological structure, including segmental and prosodic constituents, as well as their organization. The second element of W is the set M, which contains information about the verb's morphological structure. The final element of W is a set of ordered pairs, $C_{MP}$ , which defines the relationship between the phonological and morphological structure of the verb. For each pair in $C_{MP}$, the first element is a constituent in P (i.e. an element of the set $V_{phon}$, which itself is an element of the set P), and the second element is a constituent in M (i.e. an element of the set $V_{morph}$ which itself is an element of the set M). The model does not require that every vertex in $V_{phon}$ or in $V_{morph}$ belong to a pair in $C_{MP}$. The representation of the form of the English verb [stɪk] 'stick' is given in (18). This representation is an expansion of the representation given in (14) in that it includes morphological information, and information about how the phonology and morphology correspond.

(18) Expanded representation of [stɪk]

$$W = \{P,\ M,\ C_{MP}\}$$
$$P = \{U_{phon},\ V_{phon}, L_{phon}, D_{phon}, B_{phon}, N_{phon}\}$$
$$U_{phon} = \{V_{phon} \cup L_{phon}\}$$
$$V_{phon} = \{1,\ 2,\ 3,\ 4,\ 5,\ 6,\ 7,\ 8,\ 9,\ 10,\}$$
$$L_{phon} = \{'Pw',\ 'σ',\ 'O',\ 'R',\ 'N',\ 'C',\ 's',\ 't',\ 'ɪ',\ 'k'\}$$
$$D_{phon} = \{(1, 2),\ (2,\ 3),\ (2,\ 4),\ (3,\ 7),\ (3,\ 8),\ (4,\ 5),\ (4,\ 6),\ (5,\ 9),\ (6,\ 10)\}$$
$$Pr_{phon} = \{(3,\ 4),\ (5,\ 6),\ (7,\ 8)\}$$
$$N_{phon} = \{(1,\ 'Pw'),\ (2,\ 'σ'),\ (3,\ 'O'),\ (4,\ 'R'),\ (5, 'N'),\ (6,\ 'C'),\ (7,\ 's'),\ (8,\ 't'),\ (9,\ 'ɪ'),\ (10, 'k')\}$$
$$M = \{U_{morph},\ V_{morph}, L_{morph},\ D_{morph},\ B_{morph},\ N_{morph}\}$$
$$U_{morph} = \{V_{morph} \cup L_{morph}\}$$
$$V_{morph} = \{100,\ 101\}$$
$$L_{morph} = \{'Mw',\ 'Mst'\}$$
$$D_{morph} = \{(100,\ 101)\}$$
$$Pr_{morph} = \{ø\}$$
$$N_{morph} = \{(100,\ 'W'),\ (101,\ 'Mst')\}$$
$$C_{MP} = \{(1,\ 100),\ (7,\ 101),\ (8,\ 101),\ (9,\ 101),\ (10,\ 101)\}$$

The verb *stick* is an uninflected stem, therefore there are only two vertices in M–one corresponding to the stem, and one corresponding to the morphological word. All of the segments belong to the stem, so all are put into correspondence relations in $C_{MP}$ with vertex 101. The entire phonological word corresponds to the entire morphological word, therefore there is a correspondence in $C_{MP}$ between vertex 1 and vertex 101.

Regardless of whether it involves reduplication or non-reduplicative affixation, the pluractional form of verbs is derived by concatenating the form of the base verb with an affix. In the case of non-reduplicative affixation, the form of the affix is given in the input, in a set A that is set up exactly like the set W. In the case of reduplication, the form of the affix must be derived from the form of the base verb.

The first step in deriving the form of the reduplicative affix is defining the base for reduplication, which in formal terms means creating a set, Bs, that represents the information about the form of the base in the same way that W represents the information about the form of the base verb (i.e. with elements $P'$, $M'$, and $C'_{MP}$). In this model, the base for reduplication is always a morphological constituent of the base verb. The single vertex corresponding to that morphological constituent is given by applying the function **base(**x**)** to the set $V_{morph}$. The output of **base(**$V_{morph}$**)** is defined in the input. Given **base(**$V_{morph}$), the elements of the sets $U'_{morph}$, $V'_{morph}$, $L'_{morph}$, $D'_{morph}$, $Pr'_{morph}$, $N'_{morph}$ , which are the elements of the set M', can be determined according to the algorithm in (19).

(19) Algorithm for defining the morphological content of the base for reduplication (input: W)

(a) put the vertex output by **base(**$V_{morph}$**)** into the set $V'_{morph}$
(b) check each pair of vertices (x, y) in the set $D_{morph}$ and identify the ones whose first element is the same as the output of **base(**$V_{morph}$**)**
(c) for every pair of vertices identified in step (b), put the second vertex in the pair into the set $V'_{morph}$

LOOP until no new additions to $V'_{morph}$:

(d) check each pair of vertices (x, y) in the set $D_{morph}$ and identify the ones whose first element is in the set $V'_{morph}$
(e) for every pair of vertices identified in step (e), put the second vertex in the pair into the set $V'_{morph}$

END LOOP

(f) check each pair of vertices (x, y) in the set $D_{morph}$ and identify the ones where both x and y are in the set $V'_{morph}$
(g) put every pair identified in step g into the set $Pr'_{morph}$
(h) check each pair of vertices (x, y) in the set $N_{morph}$ and identify the ones whose first element is in the set $V'_{morph}$
(i) put every pair identified in step i into the set $N'_{morph}$
(j) put every label in the set $L_{morph}$ into the set $L'_{morph}$

(k) put every element that is in the set $V'_{morph}$ or the set $L'_{morph}$ into the set $U'_{morph}$

(l) output $M'$

The algorithm in (20) gives the elements of the set $C'_{MP}$–the correspondences between morphological and phonological elements in the base–, and the algorithm in (21) gives the elements of the sets $U'_{phon}$, $V'_{phon}$, $L'_{phon}$, $D'_{phon}$, $Pr'_{phon}$, $N'_{phon}$, which together make up the set $P'$–the phonological material included in the base.

(20) Algorithm for defining the morphology-phonology correspondences in the base (input: *W, M'*)

(a) check every pair of vertices (x, y) in the set $C_{MP}$ and identify the ones whose first element belongs to the set $V'_{morph}$

(b) put every pair identified in step a into the set $C'_{MP}$

(c) output $C'_{MP}$

(21) Algorithm for defining the phonological content of the base for reduplication (input: *W,* $C'_{MP}$)

(a) check every vertex in the set $V_{phon}$ and identify the ones that are the second element in a pair (x, y) in the set $C'_{MP}$

(b) put every vertex identified in step a into the set $V'_{phon}$

(c) check every pair (x, y) in the set $D_{phon}$ and identify the ones where both elements are in the set $V'_{phon}$

(d) put every pair identified in step c into the set $D'_{phon}$

(e) check every pair (x, y) in the set $Pr_{phon}$ and identify the ones where both elements are in the set $V'_{phon}$

(f) put every pair identified in step e into the set $Pr'_{phon}$

(g) put every label in the set $L_{phon}$ into the set $L'_{phon}$

(h) put every element that is in the set $V'_{phon}$ or the set $L'_{phon}$ into the set $U'_{phon}$

(i) check every pair (x, y) in the set $N_{phon}$ and identify the ones where the first element is in the set $V'_{phon}$

(j) put every pair identified in step i into the set $N'_{phon}$

(k) output $P'$

The combined output of the algorithms in (19), (20), and (21) is the completed set *Bs*. Relating this to the linguistic process being represented, we can say that applying algorithms (19), (20), and (21) to the model of a verb like in (18) will produce the base for reduplication of that verb.

After the base for reduplication has been defined, the next step is to create the reduplicant by copying material from the base into a set $Red$ that represents the form of the reduplicant in the same way that *W* represents the information about the form of the base verb (i.e. with elements $P''$, $M''$, and $C''_{MP}$). In the case of total reduplication, $P''$ (element of the set *Red*) is identical to $P'$ (element of the set *Bs*), and likewise with $M''$ and $M'$, and $C''_{MP}$ and $C'_{MP}$. Therefore, for total reduplication, creating the set *Red* involves copying every element of the base.

In the case of partial reduplication, some of the vertices (and as a result, some of the relations) that are present in the base are not part of the reduplicant. The algorithm must therefore go through extra steps in order to determine which elements of the sets in *Bs* are copied into the sets in *Red*. Which vertices (and relations) from *Bs* are included in *Red* is determined in part by the input, and in part by the algorithm. The input provides the prosodic structure of the reduplicant (as well as some segmental content in the case of reduplication patterns that have fixed segments) in the form of an incomplete set $P''$–incomplete because it is missing segmental information. The input also provides the morphological structure of the reduplicant, in the form of a set $M''$ that contains a single vertex labeled *R* (for reduplicant). The algorithm dictates how the prosodic structure given in $P''$ is filled out by segmental information contained in $P'$. Here, I show how the model derives the reduplicant of the shape Caʔ (written Ca' in the orthography). This reduplicant, used in verbal reduplication in Kaqchikel, copies the first consonant of the base and adds to it the two fixed segments in order to form a monosyllabic reduplicant. This example shows the derivation of the reduplicant *t͡saʔ* from the Kaqchikel verb *t͡six*, 'light' (Henderson, 2012).

Together, (22) and (23) define the input to the algorithm that outputs the reduplicant *t͡saʔ* for the verb *t͡six*.

(22) Set *Bs* for the Kaqchikel verb *t͡six*

(a) $Bs = \{P', M', C'_{MP}\}$

(b) $P' = \{U'_{phon}, V'_{phon}, L'_{phon}, D'_{phon}, Pr'_{phon}, N'_{phon}\}$

(c) $U'_{phon} = \{V'_{phon} \cup L'_{phon}\}$

(d) $V'_{phon} = \{0, 1, 2, 3, 4, 5, 6, 7, 8,\}$

(e) $L'_{phon} = \{'Pw', '\sigma', 'O', 'R', 'N', 'C', '\widehat{ts}', 'i', 'x'\}$

(f) $D'_{phon} = \{(0,1), (1,2), (1,3), (2,4), (3,5), (3,6), (5,7), (6,8)\}$

(g) $Pr'_{phon} = \{(2,3), (5,6)\}$

(h) $N'_{phon} = \{(0,'Pw'), (1,'\sigma'), (2,'O'), (3,'R'), (4,'\widehat{ts}'), (5,'N'), (6,'C'), (7,'i'), (8,'x')\}$

(23) Information given in the input about the set *Red* for the Kaqchikel verb *t͡six*

(a) $Red = \{P'', M'', C''_{MP}\}$

(b) $P'' = \{U''_{phon}, V''_{phon}, L''_{phon}, D''_{phon}, Pr''_{phon}, N''_{phon}\}$

(c) $U''_{phon} = \{V''_{phon} \cup L''_{phon}\}$

(d) $V''_{phon} = \{0, 1, 2, 3, 5, 6, 7, 8,\}$

(e) $L''_{phon} = \{'Pw', 'σ', 'O', 'R', 'N', 'C', 'a', 'ʔ'\}$
(f) $D''_{phon} = \{(0, 1), (1, 2), (1, 3), (3, 5), (3, 6), (5, 7), (6, 8)\}$
(g) $Pr''_{phon} = \{(2, 3), (5, 6)\}$
(h) $N''_{phon} = \{(0, 'Pw'), (1, 'σ'), (2, 'O'), (3, 'R'), (5, 'N'), (6, 'C'), (7, 'a'), (8, 'ʔ')\}$
(i) $M'' = \{U''_{morph}, V''_{morph}, L''_{morph}, D''_{morph}, Pr''_{morph}, N''_{morph}\}$
(j) $U''_{morph} = \{V''_{morph} \cup L''_{morph}\}$
(k) $V''_{morph} = \{200\}$
(l) $L''_{morph} = \{'R'\}$
(m) $D''_{morph} = \emptyset$
(n) $Pr''_{morph} = \emptyset$
(o) $N''_{morph} = \{(200, 'R')\}$
(p) $C''_{MP} = \{(0, 200), (7, 200), (8, 200)\}$

The algorithm that derives the missing elements of *Red* is given in (24). As stated above, the reduplicant is a single syllable formed from a copy of the first consonant in the base, and the fixed segments [aʔ]. The role of the algorithm is to identify the first consonant of the base, and then to put this consonant into the reduplicant, along with information about how it is ordered with respect to the fixed segments in the base. In Kaqchikel, reduplication applies to CVC roots exclusively. Therefore, the consonant that gets reduplicated can be identified by the fact that it is dominated by a vertex with the label *O* (onset). The reduplicated consonant is always the sole segment in the onset of the reduplicant, so to have it ordered correctly with respect to the fixed segments, it is sufficient to require that the vertex be dominated by a vertex with the label *O* (onset). This is shown formally in (24). Note that in this algorithm the term 'copy' is defined as the operation of creating a new element that has the same properties as the copied element but has a different name.

(24) Algorithm for creating the reduplicant for the Kaqchikel verb *t͡six* (input: *Bs*)

(a) check all of the pairs (x, y) in the set $N'_{phon}$ and copy the ones where the second element is the label 'O' into a set *On*
(b) check all of the pairs (x, y) in the set $D'_{phon}$ and for the ones where the first element belongs to *On*, copy their second element into set *Temp*
(c) copy every element in *Temp* into the set $V''_{phon}$
(d) for every element in *Temp*, create a pair (x, y) in the set $D''_{phon}$ where the first element is also the first element of a pair of the form (x, 'O') in $N''_{phon}$ and the second element is the element from *Temp*
(e) check all of the pairs (x, y) in the set $N'_{phon}$ and for the ones where the first element has a copy in the set $V''_{phon}$ put them into a set *Temp-2*

(f) for every pair in *Temp-2*, create a pair (x, y) in the set $N''_{phon}$ where y is the same as the second element of the pair from Temp-2, and x is the copy of the first element of the pair from *Temp-2*

(g) put the second element of every pair in the set $N''_{phon}$ into the set $L''_{phon}$

(h) output: *Red*

When applied to the base for reduplication for the Kaqchikel verb $\widehat{ts}ix$, (24) a-c give $4 \in V''_{phon}$, (24) d gives $(2, 4) \in D''_{phon}$, (24) e-f give (4, '$\widehat{ts}$') $\in N''_{phon}$, and (24) g. gives '$\widehat{ts}$' $\in L''_{phon}$. The completed set P" is given in (25).

(25) Output of algorithm (21): reduplicant of the Kaqchikel verb $\widehat{ts}ix$

(a) $P'' = \{U''_{phon}, V''_{phon}, L''_{phon}, D''_{phon}, Pr''_{phon}, N''_{phon}\}$

(b) $U''_{phon} = \{V''_{phon} \cup L''_{phon}\}$

(c) $V''_{phon} = \{0, 1, 2, 3, 4, 5, 6, 7, 8,\}$

(d) $L''_{phon} = \{'W', 'σ', 'O', 'R', 'N', 'C', 'a', 'ʔ', 'ts'\}$

(e) $D''_{phon} = \{(0, 1), (1, 2), (1, 3), (3, 5), (3, 6), (5, 7), (6, 8), (2, 4)\}$

(f) $Pr''_{phon} = \{(2, 3), (5, 6)\}$

(g) $N''_{phon} = \{(0, 'W'), (1, 'σ'), (2, 'O'), (3, 'R'), (5, 'N'), (6, 'C'), (7, 'a'), (8, 'ʔ'), (4, 'ts')\}$

Once the form of the affix has been either accessed from the input (in the set A) or derived, the three marking strategies–non-reduplicative affixation, total reduplication, and partial reduplication–have the same final step in creating the form of the pluractional verb. This step involves concatenating the affix with the base.

The form of the pluractional (affixed or reduplicated) verb is represented by a new set W that contains information about the phonology and morphology of the reduplicated form of the verb, along with the correspondences between the two. To distinguish the set $W$ that represents the reduplicated form from the set *W* that represents the unreduplicated form, we will label the new set $W_2$. Analogous to every previous step, $W_2$ has elements $P_2$, $M_2$, and $C_{MP\text{-}2}$.

In (26), I illustrate the concatenation process using the same Kaqchikel verb $\widehat{ts}ix$. (26) shows how the reduplicant $\widehat{ts}aʔ$ is attached as a suffix to the base $\widehat{ts}ix$. The concatenation process involves adding a vertex that dominates the entire reduplicant as well as the entire base (two dominance relations are added to encode this information), then adding a precedence relation which ensures that the affix and stem are correctly ordered. The ordering of the affix and stem is pattern-specific, so no general algorithm is given for this step. This step does not affect the complexity comparison of the three strategies because it is used in all three, and involves the same operations in each case.

(26) Output of concatenation for the Kaqchikel verb *t͡six* and affix *t͡saʔ*

(a) $W_2 = \{P_2, M_2, C_{MP-2}\}$
(b) $P_2 = \{U_{phon-2}, V_{phon-2}, L_{phon-2}, D_{phon-2}, B_{phon-2}, N_{phon-2}\}$
(c) $U_{phon-2} = \{V_{phon-2} \cup L_{phon-2}\}$
(d) $V_{phon-2} = \{V_{phon} \cup V''_{phon} \wedge 300\}$
(e) $L_{phon-2} = \{L_{phon} \cup L''_{phon}\}$
(f) $D_{phon-2} = \{D_{phon} \cup D''_{phon} \wedge (300,\ 0_{Bs}),\ (300,\ 0_{Red})$
(g) $Pr_{phon-2} = \{Pr_{phon} \cup Pr''_{phon} \wedge (0_{Bs}, 0_{Red})\}$
(h) $N_{phon-2} = \{N_{phon} \cup N''_{phon} \wedge (300, 'Pw')\}$
(i) $M_{morph-2} = \{U_{morph-2}, V_{morph-2}, L_{morph-2}, D_{morph-2}, B_{morph-2}, N_{morph-2}\}$
(j) $U_{morph-2} = \{V_{morph-2} \cup L_{morph-2}\}$
(k) $V_{morph-2} = \{V_{morph} \cup V''_{morph} \wedge 400\}$
(l) $L_{morph-2} = \{L_{morph} \cup L''_{morph}\}$
(m) $D_{morph-2} = \{D_{morph} \cup D''_{morph} \wedge (400, 100),\ (400, 200)\}$
(n) $Pr_{morph-2} = \{Pr_{morph} \cup Pr''_{morph} \wedge (100, 200)\}$
(o) $N_{morph-2} = \{N_{morph} \cup N''_{morph} \wedge (400, 'Mw')\}$
(p) $C_{MP-2} = \{C_{MP} \cup C''_{MP} \wedge (300,\ 400)\}$

Section IV.A.1 has given each of the algorithms involved in deriving the form of an affixed or reduplicated verb. The complexity of these algorithms will be considered in section IV.B. Next, section IV.A.2 gives the algorithms involved in deriving the meaning of an EP or an IP from the meaning of the base verb.

## 2. Modeling the derivation of the meaning of pluractional verbs

As introduced in section III.B and restated in (27), the meaning of a verb can be represented as follows:

(27) General model for the meaning of verbs

[[verb]] = $\{\{e_1, x_{11}, x_{12}, \ldots\}, \{e_2, x_{21}, x_{22}, \ldots\}, \ldots, \{e_m, x_{m1}, x_{m2}, \ldots\}\}$
where, for all i ∈ {1, …, m}, $e_i$ ∈ **property_v**($D_e$), and for all j ∈ {1, …, m} and k ∈ N, $x_{jk}$ ∈ **agent**($D_e$) or **theme**($D_e$). In other words, this is the set of all sets that meet the requirements for the verb.

The difference between base (non-pluractional) verbs, EPs, and IPs lies in the set of events that each type of verb picks out via its **property_v()** function. **property_v()** functions are members of the class of functions **property(). property_v()** functions take as their input a set of events and output another set of events that has the following properties: (i) It is a subset of the input set and (ii) All of the events in the set share a property determined by the verb v. Each verb has a unique property_v() function as part of its meaning, and when applied to the domain of events, this function outputs the set of all events (in the domain of events) that have the property required by the verb v (e.g. property_kiss($D_\epsilon$) function outputs a set that includes every event that have the property of being a kissing event, property_chop($D_\epsilon$) function outputs the set that includes every event that have the property of being a chopping event, etc.). In this section, I give an algorithm that derives the meaning of EPs from the meaning of the base verb, and the meaning of IPs from the base verb. In each case, the difference between the types of verbs comes from a difference in the set of events output by the **property_v()** function, so the algorithm focuses on deriving the correct set of events for the EP/IP given the set of events for the base verb.

## Definitions

**atomic(**X**)** is a blackbox function which takes as input a set of events. For example, **atomic**(**property_kiss**($D_\epsilon$)) takes as its input the set of kissing events, and outputs the set of all atomic kissing events.

**subevent_v** is a blackbox function that outputs the set of all events that language users can identify via world-knowledge as being implied by the type of event identified by the **property_v()** function for the same verb. For example, a language user may code puckering the lips as the only subevent that is implied to have taken place whenever there is a kissing event. With this way of encoding events, **subevent_kiss** would output the set of lip-puckering events.

**SPS_v(**Y**)** is a class of blackbox functions with a unique member corresponding to each verb v in the language. Functions in the **SPS_v(**Y**)** class take as their input a set of events Y, and then output the set of all atomic events in Y that are spatiotemporally superimposed over a repetition of an event of the type identified by the **property_v()** function for the verb v. For example, **SPS_kiss**($D_\epsilon$) checks the entire domain of events and identifies atomic events that are spatiotemporally superimposed on a repetition of kissing events.

## Deriving the meaning of pluractionals

Recall that under Henderson's (2012) analysis, EPs are satisfied by non-atomic events with the property specified by the base predicate. The requirement that events be non-atomic excludes single-event scenarios as well as events that satisfy the requirements of IPs rather than EPs. The algorithm in (28) takes the set of events that satisfy a base predicate (the powerset of an **atomicproperty_v()** function) and removes all of the atomic events to output the set of events that satisfy the EP.

(28) Algorithm for the meaning of EPs (input: property_v($D_\epsilon$) – the set of events that satisfy the base verb)

(a) get the powerset of **atomic(property_v**($D_\epsilon$)) and place every element in the set *Temp*
(b) check every element in *Temp* and place the ones that are not also in **atomic(property_v**($D_\epsilon$)) into the set *EP_meaning*
(c) output: *EP_meaning*

In Henderson's (2012) proposal, IPs can only be satisfied by atomic events. However, not all of the events returned by **atomic(property_v**($D_\epsilon$)) can satisfy an IP: only events that are spatiotemporally superimposed over a repetition of (a subphase of) an event of the specified by the **property_v**() function. Because the subphase events are not included in the output of the **property_v()** function, the entire domain of events ($D_\epsilon$) must be included in the input. The algorithm in (29) takes the set of atomic events and picks out those that meet the plurality criteria in order to get the set of events that satisfy the IP.

(29) Algorithm for the meaning of EPs (input: property_v($D_\epsilon$ ), $D_\epsilon$ – the set of events that satisfy the base verb, and the domain of events)

(a) Apply **SPS_v()** to **atomic(property_v**($D_\epsilon$)) and place every element output into the set *IP_meaning*
(b) For verb u in **subevent_v** :
    Add **SPS_u**($D_\epsilon$) to the set *IP_meaning*
(c) output: *IP_meaning*

In section IV.A, we have defined algorithms that represent the four processes of interest in this paper: reduplication, non-reduplicative affixation, deriving the meaning of IPs, and deriving the meaning of EPs. These algorithms take set-theoretic models of the forms or meanings of non-pluractional verbs as their inputs, and give set-theoretic models of the forms or meanings of pluractional verbs as their outputs.

In order to determine whether the pattern observed in Kaqchikel, Karuk, and Yurok is iconic along the dimension of complexity, we need to determine the relative complexity of the two marking strategies (i.e. determine whether reduplication or non-reduplicative affixation is more complex), and determine the relative complexity of the two types of pluractionality (i.e. determine whether IPs or EPs are more complex). The final step, then, is to check whether the more complex marking strategy is paired with the more complex type of pluractionality. If yes, then the property of increased (relative) complexity is shared by the form and the meaning, and so we can say that the pattern is iconic along the dimension of complexity. In the following section, we calculate the relative complexity of the linguistic processes, and ultimately conclude that the pattern of marking IPs with reduplication and EPs with non-reduplicative affixation is iconic.

## B. Algorithm analysis

In this section, we determine the computational complexity of affixation, total reduplication, partial reduplication, derivation of EPs, and derivation of IPs. In order to find the complexity of the process, we first identify the algorithms used in the process, then calculate the complexity of each algorithm based on the results in Knuth (1997). The overall complexity of each process is equal to the complexity of the most complex algorithm that the process uses. Two algorithms are distinct in terms of their complexity if the complexity of one belongs to a different order than the complexity of the other (e.g. n vs. $n^2$ vs. $n^3$…).

The measure of computational complexity that I am using can be conceptualized in terms of the number of steps that an algorithm requires in order to complete its task. This is dependent on (i) the type of operations that the algorithm must perform, and (ii) the size of the input to the algorithm. As an example of how (i) affects complexity, compare an operation that requires a single person to ask the name of every person in their department with an operation that requires every person to ask the name of every other person in the department. If there are 20 people in the department then the former operation would require 20 questions (assuming for simplicity that the operation requires the person to ask their own name as well), while the latter operation would require each of the 20 people to ask 20 questions, i.e. $20^2$ questions. This translates to a complexity of O(n) vs. a complexity of O($n^2$) where n is the input size (size of the department). To understand the contribution of (ii), compare a situation where every person in the university has to ask the name of every other person in the university. If there are even 400 people in this university then the operation would require 400 people to ask 400 questions each, resulting in $20^4$ questions altogether. In this example, performing the same operation that yielded a complexity of O($n^2$) with the smaller input yields a complexity of O($n^4$) with the larger input.

Furthermore, computational complexity is characterized by its highest order. That is, if a person's task is to ask the name of every person in their department then do a dance, then the amount of operations they would need to perform is O($n^2$ + 1). However, because the first term, $n^2$, is of a higher order than the second term, 1, the expression ($n^2$ + 1) is dominated by the first term. In other words, the value of this expression ($n^2$ + 1) almost solely depends on the value of the first term (especially at large input sizes). Since computational complexity generally works with input sizes of extreme magnitudes (in the order of thousands, hundreds of thousands, and millions), we reduce computational complexity expressions down to its term of the highest order. Therefore, a task that requires O($n^2$ + 1) operations is said to have complexity O($n^2$).

## 1. Complexity comparison for affixation, total reduplication, and partial reduplication

Affixation is accomplished through concatenation, which does not have a generalized algorithm because determining where the affix should attach to the stem is language specific. Regardless of the pattern, concatenation involves the operation of inserting elements into sets. Total reduplication is accomplished with algorithms (19), (20), (21) and concatenation. Partial reduplication is accomplished with algorithms (19), (20), (21), (24), and concatenation.

**Theorem 1**: Partial reduplication has complexity O($V^3$)

**Proof**:

For 19: Putting the vertex output by **base($V_{morph}$)** into the set $V'_{morph}$ takes **O(1)** time. Checking each pair of vertices in the set $D_{morph}$ and identifying the ones whose first element is the same as the output of

**base($V_{morph}$)** takes **O(2D)** time, where D is the size of the set $D_{morph}$. Putting the second vertex of every pair of vertices identified in the previous step into the set $V'_{morph}$ takes **O(D)** time, where D is the size of the set $D_{morph}$. Checking each pair of vertices in the set $D_{morph}$ and identifying the ones whose first element is in the set $V'_{morph}$ takes **O(DV)** time, where D is the size of the set $D_{morph}$ and V is the size of the set $V'_{morph}$. Placing the second vertex of every pair of vertices identified in the previous step into the set $V'_{morph}$ takes **O(D)** time, where D is the size of the set $D_{morph}$. Looping the previous two steps increases their complexity to **O($D^2$V)** and **O($D^2$)** respectively. Checking each pair of vertices in the set $D_{morph}$ and identifying the ones where both x and y are in the set $V'_{morph}$ takes time **O(2DV)**, where D is the size of the set $D_{morph}$ and V is the size of the set $V'_{morph}$. Putting every pair identified in the previous step into the set $Pr'_{morph}$ takes time **O(D)**, where D is the size of the set $D_{morph}$. Check each pair of vertices in the set $N_{morph}$ and identifying the ones whose first element is in the set $V'_{morph}$ takes **O(NV)** time, where N is the size of $N_{morph}$ and V is the size of $V'_{morph}$. Put every pair identified in the previous step into the set $N'_{morph}$ takes **O(N)** time, where N is the size of $N'_{morph}$. Putting every label in the set $L_{morph}$ into the set $L'_{morph}$ takes **O(L)** time, where L is the size of $L_{morph}$. Putting every element that is in the set $V'_{morph}$ or the set $L'_{morph}$ into the set $U'_{morph}$ takes **O(V+L)** time, where V is the size of $V'_{morph}$ and L is the size of $L'_{morph}$. Therefore, the complexity of (19) is 1 + 2D + D + D + D(DV + D) + 2DV + D + NV + N + L + V + L. Without loss of generality, assume that V ≥ D, N, L. Then, the complexity of (19) is 1 + 5V + $V^3$ + $V^2$ + 2$V^2$ + $V^2$ + V + 2V + V. **Therefore, the complexity of (19) is O($V^3$).**

For 20: Checking every pair of vertices in the set $C_{MP}$ and identifying the ones whose first element belongs to the set $V'_{morph}$ takes **O(CV)** time, where C is the size of $C_{MP}$ and V is the size of $V'_{morph}$. Putting every pair identified in the previous step into the set $C'_{MP}$ takes **O(C)** time, where C is the size of $C'_{MP}$ . Therefore, the complexity of (20) is CV+C. Without loss of generality, assume that V ≥ C. Then, the complexity of (20) is $V^2$+V. **Therefore the complexity of (41) is O($V^2$).**

For 21: Checking every vertex in the set $V_{phon}$ and identifying the ones that are the second element in a pair in the set $C'_{MP}$ takes **O(V'C')** time, where V' is the size of $V_{phon}$ and **C'** is the size of $C'_{MP}$. Putting every vertex identified in the previous step into the set $V'_{phon}$ takes **O(V')** times, where V' is the size of $V_{phon}$. Checking every pair in the set $D_{phon}$ and identifying the ones where both elements are in the set $V'_{phon}$ takes **O(2D'V")** time, where D' is the size of $D_{phon}$ and V" is the size of $V'_{phon}$ . Putting every pair identified in the previous step into the set $D'_{phon}$ takes **O(D')** time, where D' is the size of $D_{phon}$. Checking every pair in the set $Pr_{phon}$ and identifying the ones where both elements are in the set $V'_{phon}$ takes **O(RV")** time where R is the size of $Pr_{phon}$ and V" is the size of $V'_{phon}$. Putting every pair identified in the previous step into the set $Pr'_{phon}$ takes **O(R)** time, where R is the size of $Pr_{phon}$. Putting every label in the set $L_{phon}$ into the set $L'_{phon}$ takes **O(L')** time where L' is the size of $L_{phon}$. Putting every element that is in the set $V'_{phon}$ or the set $L'_{phon}$ into the set $U'_{phon}$ takes **O(V"+L')** time where V" is the size of $V'_{phon}$ and L' is the size of $L'_{phon}$. Checking every pair in the set $N_{phon}$ and identifying the ones where the first element is in the set $V'_{phon}$ takes **O(N'V")** where N' is the size of $N_{phon}$ and V" is the size of $V'_{phon}$. Putting every pair identified in the previous step into the set $N'_{phon}$ takes **O(N')** time, where N' is the size of $N_{phon}$. Therefore, the complexity of (21) is O(C'V'+V'+2D'V"+D'+RV"+R+L'+V"+L'+N'+V"). Without loss of generality, assume that V' ≥ C', D', V", R, L', N'. Then, the complexity of (18) is $O(V'^2+V'+2V'^2+V'+V'^2+V'+V'+V'+V'+V'+V')$. **Therefore, the complexity of (21) is $O(V'^2)$.**

For 24: Checking all of the pairs in the set $N'_{phon}$ and copying the ones where the second element is the label 'O' into a set *On* takes **O(N")** time where N" is the size of $N'_{phon}$. Checking all of the pairs (x, y) in the set $D'_{phon}$ and for the ones where the first element belongs to *On*, copying their second element into set *Temp* takes **O(2D")** time, where D" is the size of $D'_{phon}$. Copying every element in *Temp* into the set $V''_{phon}$ takes **O(D")** time, where D" is the size of $D'_{phon}$. Creating a pair in the set $D''_{phon}$ for every element in *Temp* where the first element is also the first element of a pair of the form (x, 'O') in $N''_{phon}$ and the second element is the element from *Temp* takes **O(2D"+N"')** time where D" is the size of $D'_{phon}$ and N"' is the size of $N''_{phon}$. Checking all of the pairs in the set $N'_{phon}$ and for the ones where the first element has a copy in the set $V''_{phon}$ putting them into a set *Temp-2* takes **O(N"V"'+N")** time where N" is the size of $N'_{phon}$ and **V"'** is the size of $V''_{phon}$. Creating a pair (x, y) in $N''_{phon}$ for every pair in *Temp-2* where y is the same as the second element of the pair from *Temp-2*, and x is the copy of the first element of the pair from *Temp-2* takes **O(2N")** time, where N"is the size of $N'_{phon}$. Putting the second element of every pair in the set $N''_{phon}$ into the set $L''_{phon}$ takes **O(N"')** time, where N"' is the size of $N''_{phon}$. Therefore, the complexity of (24) is O(N"+2D"+D"+2D"+N"'+N"V"'+N"+2N"). Without loss of generality, assume that V"' ≥ N", D", N", N"'. Then, complexity of (24) is $O(V''+2V'''+V'''+2V'''+V'''+V'''^2+V''+2V'')$. **Therefore, the complexity of (24) is $O(V'''^2)$.**

For concatenation: **Assume lowest complexity, O(1)**.

Without loss of generality, assume that V ≥ V', V"', **Therefore, the complexity of partial reduplication is $O(V^3)$.**

**Theorem 2:** Total reduplication has complexity $O(V^3)$

**Proof:**

Complexity of (19) is $O(V^3)$. Complexity of (20) is $O(V^2)$. Complexity of (21) is $O(V'^2)$. The complexity of concatenation is O(1). Without loss of generality, assume that V ≥ V', V"', **Therefore, the complexity of total reduplication is $O(V^3)$.**

**Theorem 3:** affixation has complexity O(1)

**Proof:**

The complexity of concatenation is O(1). Therefore, the complexity of affixation is **O(1)**.

**Based on the above analysis, we conclude that non-repduplicative affixation** (complexity = O(1)) **is less complex than reduplication** (complexity = $O(V^3)$. In the following section, we conduct a similar analysis to determine the complexity of IPs and EPs in order to ultimately verify whether reduplication, which is the more complex marking strategy, is paired with the more complex meaning.

## 2. Complexity comparison for IPs and EPs

The meaning of EPs is obtained with algorithm (28), and the meaning of IPs is obtained with algorithm (29).

**Theorem 4:** the complexity of calculating the meaning of an EP is $O(2^A)$.

**Proof:**

Getting the powerset of **atomic(property_v**($D_\epsilon$)) and place every element in the set *Temp* takes $\mathbf{O(2^A)}$ time, where A is the size of **atomic(property_v**($D_\epsilon$)). Checking every element in *Temp* and place the ones that are not also in **atomic(property_v**($D_\epsilon$)) into the set *EP_meaning* takes $\mathbf{O(A^2)}$ time, where A is the size of **atomic(property_v**($D_\epsilon$)). Therefore, the complexity of (28) is $O(2^A+A^2)$. Without loss of generality, assume $A \geq 2$. **Therefore, the complexity of (28) is $\mathbf{O(2^A)}$.**

**Theorem 5:** the complexity of calculating the meaning of an IP is $O(E^2)$

**Proof:**

Applying **SPS_v()** to **atomic(property_v**($D_\epsilon$)) and placing every element output into the set *IP_meaning* takes **O(A)** time, where A is the size of **atomic(property_v**($D_\epsilon$)). Adding **SPS_u**($D_\epsilon$) to the set *IP_meaning* for every verb u identified by **subevent_v** takes $\mathbf{O(E^2)}$ time, where E is the size of the domain of events. Therefore, the complexity of (29) is $O(A+E^2)$. Without loss of generality, assume $E \geq A$. **Therefore, the complexity of (29) is $\mathbf{O(E^2)}$.**

**Lemma 1:** $E \geq 2^A$

**Proof:**

A is the size of the set **atomic(property_v**($D_\epsilon$)) of a verb v. Otherwise put, A is the number of atomic events with the property dictated by a verb v. $2^A$ is equal to the size of the powerset of A. The powerset of A is the set of all (atomic or non-atomic) events with the property dictated by a verb v. The set of all events that have the property dictated by a verb v is always a subset of the domain of events (the set of all events), which has size E. **Therefore, $E \geq 2^A$** .

**Theorem 7:** calculating the meaning of IPs requires an algorithm of a higher order of complexity than the algorithm required for calculating the meaning of EPs.

Proof: By Theorem 5 and 6, the complexity of calculating the meaning of an EP is $O(2^A)$ and the complexity of calculating the meaning of an IP is $O(E^2)$. A is the size of the set of atomic events with the property specified by a verb v, and E is the size of the domain of events. By Lemma 6, $E \geq 2^A$. Therefore, the complexity of calculating the meaning of an EP is **no greater** than O(e), while the complexity of calculating the meaning of an IP is $O(E^2)$. **Therefore, calculating the meaning of an IP is more complex than calculating the meaning of an EP.**

## 3. Summary

We have modeled the forms and meanings of verbs as sets, and have defined algorithms that perform the following four processes:

(30) Processes captured by §IV algorithms

(a) Produce a model for the reduplicated form of a verb given the model for the unreduplicated form of the verb (reduplication)

(b) Produce a model for the affixed form of a verb given the model for the affixed form of the verb (non-reduplicative affixation)

(c) Produce a model for the meaning of an IP given the model for the meaning of the base predicate (calculating the meaning of IPs)

(d) Produce a model for the meaning of an EP given the model for the meaning of the base predicate (calculating the meaning of EPs)

The algorithm analysis given in section 4 shows that (30.1) is more complex than (30.2), and (30.3) is more complex than (30.4). These results are significant for two reasons. Firstly, they show that complexity is a property that can distinguish between both related forms and related meanings. This means that complexity can theoretically be encoded in the grammar of a language as a dimension of iconicity. Secondly, because Kaqchikel, Karuk, and Yurok all pair reduplication (more complex) with IPs (more complex), and non-reduplicative affixation (less complex) with EPs (less complex), the pattern found in these three languages can be understood as being iconic rather than arbitrary.

# Discussion

The purpose of this paper was to provide a proof-of-concept for the idea that model-theoretic approaches to language can help to revel linguistic organization that exists below the surface. We have used pluractional marking in Karuk, Yurok, and Kaqchikel as a case study, showing that although we cannot explain the pattern found in these three languages in terms of the dimensions of iconicity that work for other patterns (e.g. the dimension of sequentiality, which can explain part of the Balinese pluractional system), we can understand the pattern as being iconic when we look at it from a formal perspective.

Our work illustrates one way that model theoretic approaches can shed light on natural langauage phenomena, but it is important to note that there are different ways of measuring complexity, and we do not make the claim that the measure chosen here is the one that aligns most closely with cognitive complexity. We leave the task of determining how best to capture and formalize cognitive complexity to future work.